\documentclass[10pt]{article} 
\usepackage[accepted]{rlc}

\usepackage{amssymb}            
\usepackage{mathtools}          
\usepackage{mathrsfs}           
\usepackage{graphicx}           
\usepackage{subcaption}         
\usepackage[space]{grffile}     
\usepackage{url}            
\usepackage{booktabs}       
\usepackage{amsfonts}       
\usepackage{nicefrac}       
\usepackage{microtype}      
\usepackage{xcolor}         
\usepackage{graphicx}
\usepackage{subcaption}
\usepackage{wrapfig}
\usepackage{lipsum}
\usepackage[linesnumbered,ruled,vlined]{algorithm2e}

\SetCommentSty{mycommfont}

\usepackage{amsmath}
\DeclareMathOperator*{\argmax}{arg\,max}
\DeclareMathOperator*{\argmin}{arg\,min}

\SetKwInput{KwInput}{Input}                
\SetKwInput{KwOutput}{Output} 
\title{Learning Rate-Free Reinforcement Learning: A Case for Model Selection with Non-Stationary Objectives}

\author{Aida Afshar\\
  Boston University\\
  \texttt{aafshar@bu.edu} \\
  \AND 
  Aldo Pacchiano\\
  Boston University \\
  Broad Institute of MIT and Harvard\\
  \texttt{pacchian@bu.edu} \\
}

\begin{document}

\maketitle

\begin{abstract}
The performance of reinforcement learning (RL) algorithms is sensitive to the choice of hyperparameters, with the learning rate being particularly influential. RL algorithms fail to reach convergence or demand an extensive number of samples when the learning rate is not optimally set. In this work, we show that model selection can help to improve the failure modes of RL that are due to suboptimal choices of learning rate. We present a model selection framework for Learning Rate-Free Reinforcement Learning that employs model selection methods to select the optimal learning rate on the fly. This approach of adaptive learning rate tuning neither depends on the underlying RL algorithm nor the optimizer and solely uses the reward feedback to select the learning rate; hence, the framework can input any RL algorithm and produce a learning rate-free version of it. We conduct experiments for policy optimization methods and evaluate various model selection strategies within our framework. Our results indicate that data-driven model selection algorithms are better alternatives to standard bandit algorithms when the optimal choice of hyperparameter is time-dependent and non-stationary.

\end{abstract}

\section{Introduction}
We focus on sequential decision-making problems such as reinforcement learning and bandits \citep{lattimore2020bandit, sutton2018reinforcement} where a learner interacts with the world in a sequential manner and is tasked with finding a policy that maximizes the reward. It is a common scenario in RL that the right choice of hyperparameters is not known in advance and the success of RL algorithms demands the effort of hyperparameter tuning \citep{eimer2023hyperparameters} to reach convergence. Among all the algorithm-specific hyperparameters, the learning rate is known to have a notable impact on the convergence of RL agents. The learning rate determines the extent to which the model parameters are adjusted at each optimization step, and prior literature on learning rate scheduling \citep{mishchenko2023prodigy, defazio2023and} suggests that optimal learning rate is dependent on the \emph{distance to the solution}. Central to RL, is the notion of reward that contains information about the proximity of the current policy to the optimal one. Building on this intuition that the reward feedback can be used as a proxy of the distance to the solution, we propose a framework that utilizes the empirical reward to adjust the learning rate on the fly during RL training. Model selection methods are inherently designed with the goal of detecting the right configuration of the problem and are uniquely suitable for the task of hyperparameter selection in RL;

\begin{enumerate}

\item Model selection algorithms advance the state of each hyperparameter curve adaptively, thus not requiring the same samples and compute for all hyperparameter choices. We show that by regret balancing, model selection will not select an ill-performing learning rate for more than $\sqrt{N}$ episodes in a single run, where $N$ is the total number of episodes. 

\item Standard methods for hyperparameter tuning such as Bayesian Optimization \citep{wu2019hyperparameter} or Random Search \citep{bergstra2012random} require multiple training runs to indicate a good choice of hyperparameter, and are not adaptive to the non-stationarity of many RL scenarios. Model Selection methods are designed to select the best-performing configuration in a single run, and we show that data-driven model selection algorithms are able to adapt to the non-stationarity of the objective hyperparameter throughout the training.

\end{enumerate}

In this paper, we show that model model selection methods can be simply integrated into RL algorithms without heavily changing the original implementation. Given a set of learning rates, the model selection method can select the optimal learning rate on the fly, removing the manual effort of learning rate tuning in RL. We evaluate six model selection strategies in this framework including Regret Balancing methods \citep{dann2024data, pacchiano2020regret}, Corraling \citep{agarwal2017corralling}, and standard bandit strategies like UCB and EXP3 \citep{auer2002finite, bubeck2012regret}. Our results suggest that data-driven regret balancing methods achieve the best performance for the task of learning rate selection in RL. Additionally, we show that when the optimal choice of learning rate is time-dependent and varies with the state of learning, bandit strategies like UCB and EXP3 are not sufficient for the task of learning rate selection. \footnote{Codes are available at \url{https://github.com/AidaAfshar/Learning-Rate-Free-Reinforcement-Learning}.}

\section{Preliminaries and Background}
\label{sec:preliminaries}
In many machine learning domains, including reinforcement learning the true configuration of the problem is not known in advance. The goal of model selection is to consider several configurations and add a strategy on top that learns to pick up the best configuration adaptively. We call each configuration a base and refer to the model selection strategy as the meta-learner. The meta-learner has access to a set of $m$ bases, in this case, different copies of the same reinforcement learning algorithm instantiated with different learning rates. In each round, $n = 1, 2, \ldots, N$, of the interaction between the meta-learner with the environment, the meta-learner selects a base $i_n \in [m]$ to play and follows its policy. Base $i_n$'s internal state is then updated with the data collected from its interaction with the environment. 

We review a few definitions from bandits to better explain these methods. The \emph{regret} of a policy $\pi$ is defined as 
\begin{equation}
   reg(\pi) = v^{\star} - v^{\pi}, 
\end{equation}

where $v^{\star}$ is the value of the optimal policy $\pi^{\star} \in \argmax_{\pi}{v^{\pi}}$. In stochastic settings, it's most common to assume that the total regret of base $i$ after being played for $k$ rounds is bounded by $\sqrt{k}$ rate,
\begin{equation}
   \sum_{l=1}^{k} reg(\pi^i_{(l)}) \leq d^i_{(k)} \sqrt{k},
\end{equation}
where $d^i_{(k)}$ is called the \emph{regret coefficient} of base $i$. The subscripts $(k)$ denote the number of times that a base has been played up to this round.

Regret balancing methods aim to equate the regret bounds across all the bases. In this approach, the base agent is selected for two reasons. It is either a well-performing base by achieving low regret, or it has not been played enough and the meta-learner hasn't collected adequate information on the performance of this base. Here, we investigate Doubling Data Driven Regret Balancing (D$^3$RB), and Estimating Data-Driven Regret Balancing (ED$^2$RB) \citep{dann2024data}. D$^3$RB maintains an estimate of the regret coefficient of base $i$ and performs a miss-specification test to see whether this estimate is compatible with data that has been collected so far. When base $i$ is selected for the $k$'th time, if the estimate is too small to correctly represent the regret of base $i$, it doubles the estimated regret coefficient $d^i_{(k)}$. On the other hand, ED$^2$RB directly estimates $d^i_{(k)}$ as the maximum difference of average reward between base $i$ and other bases, scales by $\sqrt{k}$. Both methods, update the regret bound for the selected base and then choose the base with the lowest regret for the next round. These methods are designed to work with the \emph{realized} regret as opposed to the \emph{expected} regret which is used in other model selection methods like Regret Bound Balancing \citep{pacchiano2020regret}, and Corral \citep{agarwal2017corralling, pacchiano2020model}.

The Regret Bound Balancing algorithm which we will refer to as "Classic Balancing" takes as input \emph{fixed} regret coefficients that it uses to perform a miss-specification test. When a base learner fails this test the Classic Balancing algorithm eliminates it. This makes it crucial to strike an appropriate balance when deciding what putative regret coefficients are used to initialize the meta learner. When these input coefficients underestimate the real ones, Classic Balancing may eliminate the optimal base learner. If instead, the input coefficients overestimate the true regret coefficients, Classic Balancing may over-exploit sub-optimal base learners. Corral~\citep{agarwal2017corralling} is a model selection meta-learning algorithm based on a carefully designed adversarial bandit strategy that uses log barrier regularization. Unfortunately, the regret guarantees the Corral algorithm satisfies are only known to hold in expectation, and not in high probability. We also investigate the Upper Confidence Bound (UCB) \citep{auer2002finite}, the Exponential-weight algorithm for Exploration and Exploitation (EXP3) \citep{bubeck2012regret} which has been recommended by the previous literature \citep{li2018hyperband} for hyperparameter tuning. The theoretical details and pseudo-code of all six algorithms are included in \hyperref[AppendixC]{Appendix C}. 

\section{Method}
We formalize the model selection framework for learning rate-free reinforcement learning as the tuple $\langle m, {\beta}, M, \Psi \rangle$ where $m$ is the number of base agents, $\beta = \{\beta^1, \ldots, \beta^m \}$ denotes the set of base agents where  $\beta^i =  \langle \alpha^i, \pi^i \rangle$   $(1\leq i \leq m)$ consists of learning rate $\alpha^i$, and policy $\pi^i$. Lastly, $M$ is the model selection strategy and $\Psi$ is an attribute of $M$ that expresses some statistics over the base agents. For instance, $\Psi$ can either be a distribution $\Psi: \beta \rightarrow P(\beta)$ over base agents or represent the estimated empirical regret of the base agents. 

\begin{algorithm}[H]
\SetKwInput{KwInput}{Input}                
\SetKwInput{KwOutput}{Output}              
\DontPrintSemicolon
  
  \KwInput{$m$, $\beta$, $\Psi$, $M$}

  \SetKwFunction{FMain}{Main}
  \SetKwFunction{FSum}{sample}
  \SetKwFunction{FSub}{update}
 
  \SetKwProg{Fn}{Function}{:}{}
  \Fn{\FSum{}}{
        \tcp{Select base index according to $\Psi$}
        $i \sim \Psi$\;
        $\pi^i, \alpha^i \leftarrow \beta^i$ \;
        \KwRet $i, \pi^{i}, \alpha^{i}$\;
  }
  \SetKwProg{Fn}{Function}{:}{}
  \Fn{\FSub{$index$, $R[1:T]$}}{
    \tcp{Calculate and normalize the episodic return}
        $R_{norm} \leftarrow normalize (R[1:T])$ \;
    \tcp{Update base statistics according to the meta learning algorithm $M$}
        $\Psi \leftarrow M(\Psi, index, R_{norm})$ \;
  }
\label{alg1}
\caption{Model Selection Interface for Hyperparameter Tuning}
\end{algorithm}

At the beginning of each episode, the meta learner $M$ selects base agent $\beta^j$ according to $\Psi$. We abbreviate this as $j \sim \Psi$. The base agent interacts with the environment in a typical reinforcement learning manner for one episode. At state $s_t \in S$, the base agent takes action $a_t \sim \pi^j,$  receives reward $r_t \in (0, 1)$, and move to the next state $s_{t+1} \in S$ following the environment transition dynamics. At the end of each episode, the base agent passes the realized rewards $(r_1, \ldots, r_T)$ to the meta learner, so that it updates $\Psi$ based on the model selection strategy $M$.

The goal of the base agents is to interact with the environment and learn an optimal policy for the reinforcement learning problem. The goal of the meta learner is to learn a strategy to iteratively select base agents, so that agents with better learning rates are played more frequently. It's unique to model selection that neither the base agents with good learning rates nor the optimal reinforcement strategy are known in advance, and the framework learns both of them during a single run. The model selection interface, represented in \hyperref[alg1]{Algorithm 1}, consists of two procedures $sample$, and $update$ that the meta learner uses to select the base agent at the beginning of each episode, and update $\Psi$ at the end of it. The protocol to integrate the model selection interface with the reinforcement learning loop is represented in \hyperref[alg2]{Algorithm 2} of \hyperref[AppendixA]{Appendix A}.

\section{Experiments and Results}

\begin{figure}[htb!]
    \centering 
    \includegraphics[width=0.6\textwidth]{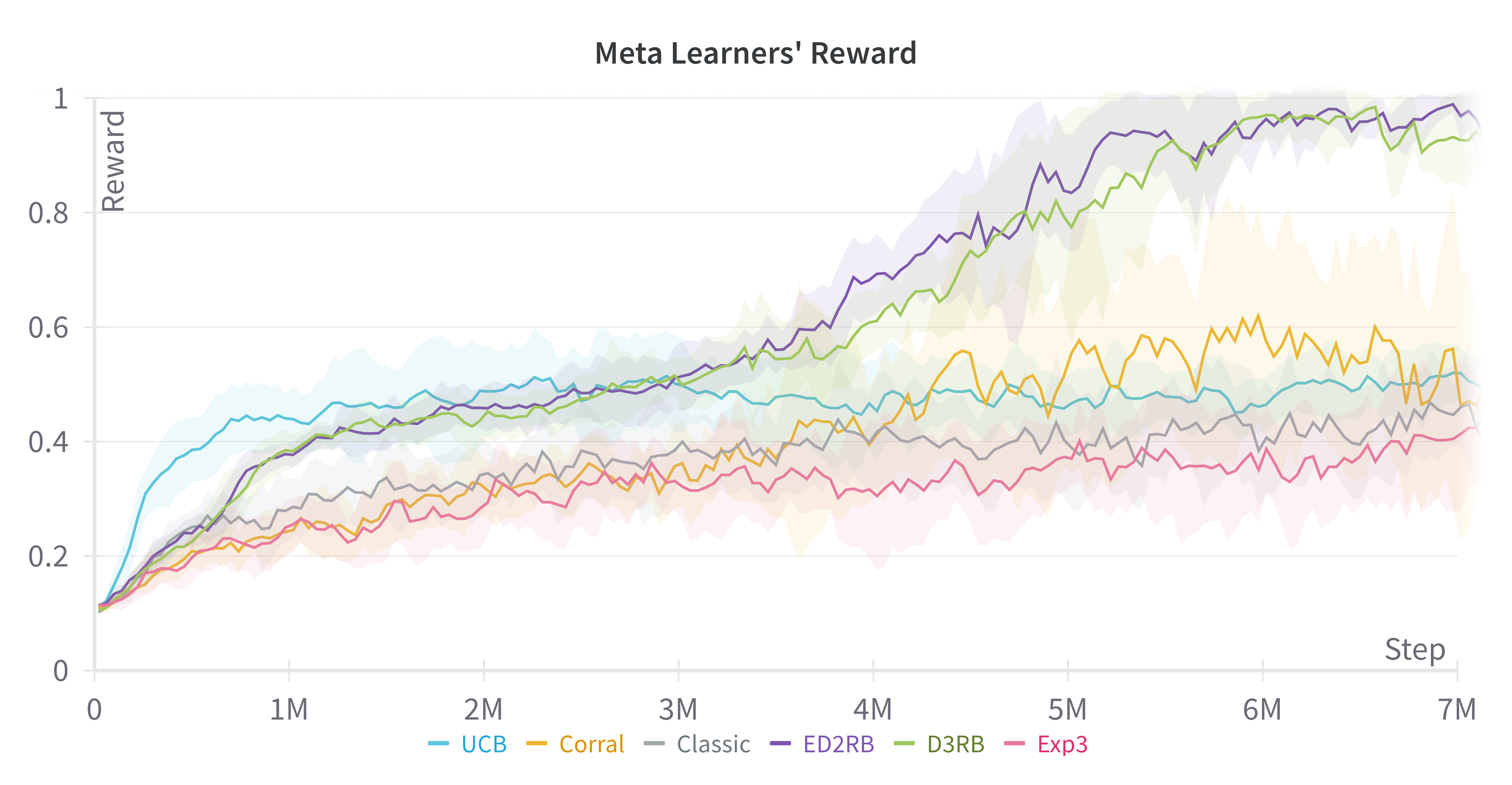}
    
    \caption{Learning Rate-Free PPO on Humanoid Environment. Each curve shows the mean and standard deviation of normalized reward per step over three seeds.}
    
    \label{fig:fig1}
\end{figure}

We begin our experiments with learning rate-free PPO. We initiate ten PPO base agents learning rates $\alpha = [1e^{-2}, 5e^{-3}, 1e^{-3}, 5e^{-4}, 1e^{-4}, 5e^{-5}, 1e^{-5}, 5e^{-6}, 1e^{-6}, 5e^{-7}]$. We run the experiment for six model selection strategies introduced in \hyperref[preliminaries]{Section 2}. \hyperref[fig:fig1]{Figure 1} represents the reward plot of six meta learners on the Mujoco Humanoid environment \citep{6386025}. By comparing the meta learners, we can see that D$^3$RB and ED$^2$RB strategies achieved the lowest regret and had the most advancement in the reward curve. The meta learners with Corral and Classic Balancing strategies, in addition to MAB meta learners, are showing sub-optimal performance in this task.

\begin{figure}[htb]
    \centering 

\begin{subfigure}{0.33\textwidth}
  \includegraphics[width=\textwidth]{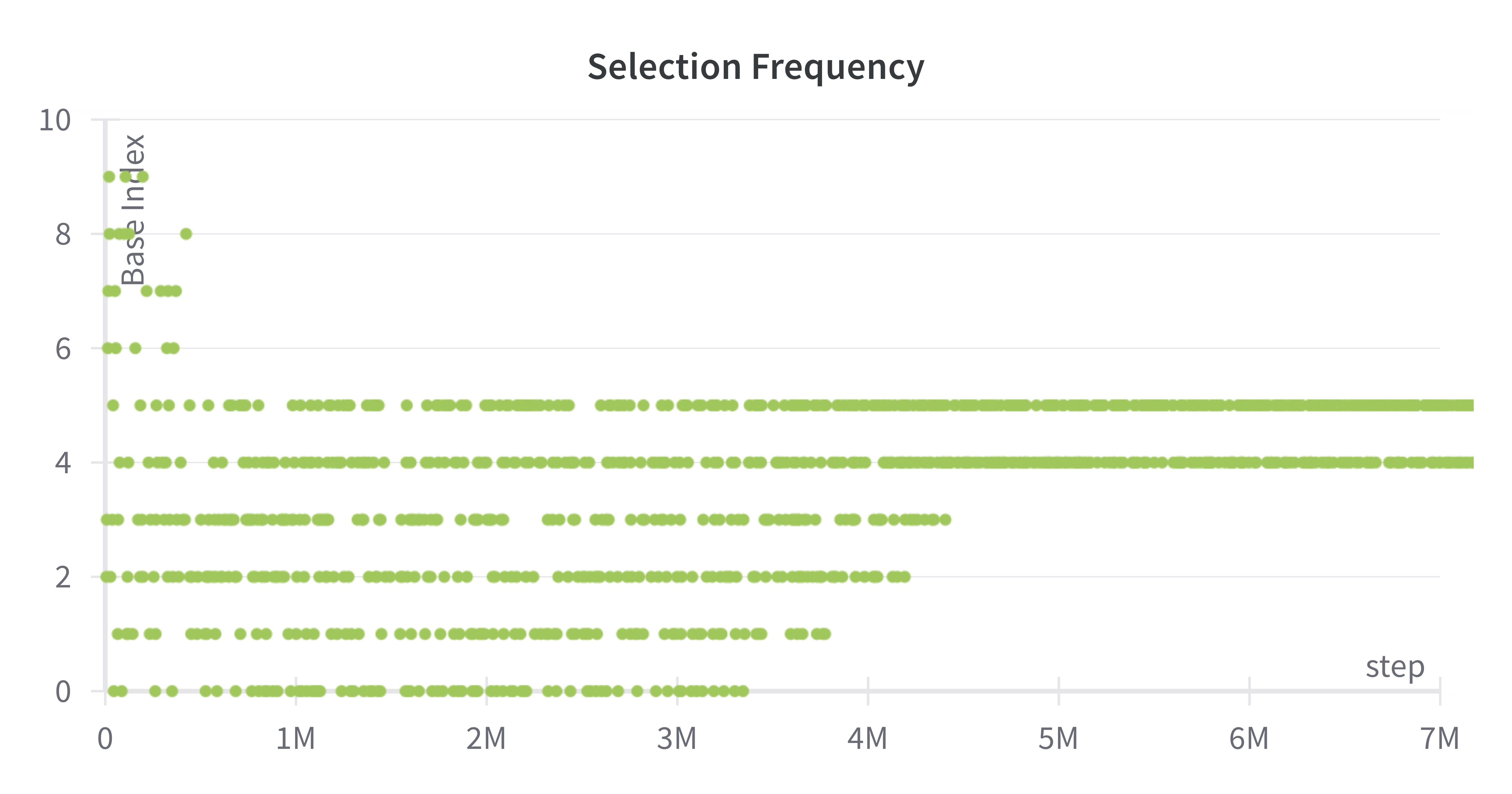}
  \caption{D$^3$RB}
  \label{fig:3}
\end{subfigure}\hfil 
\begin{subfigure}{0.33\textwidth}
  \includegraphics[width=\textwidth]{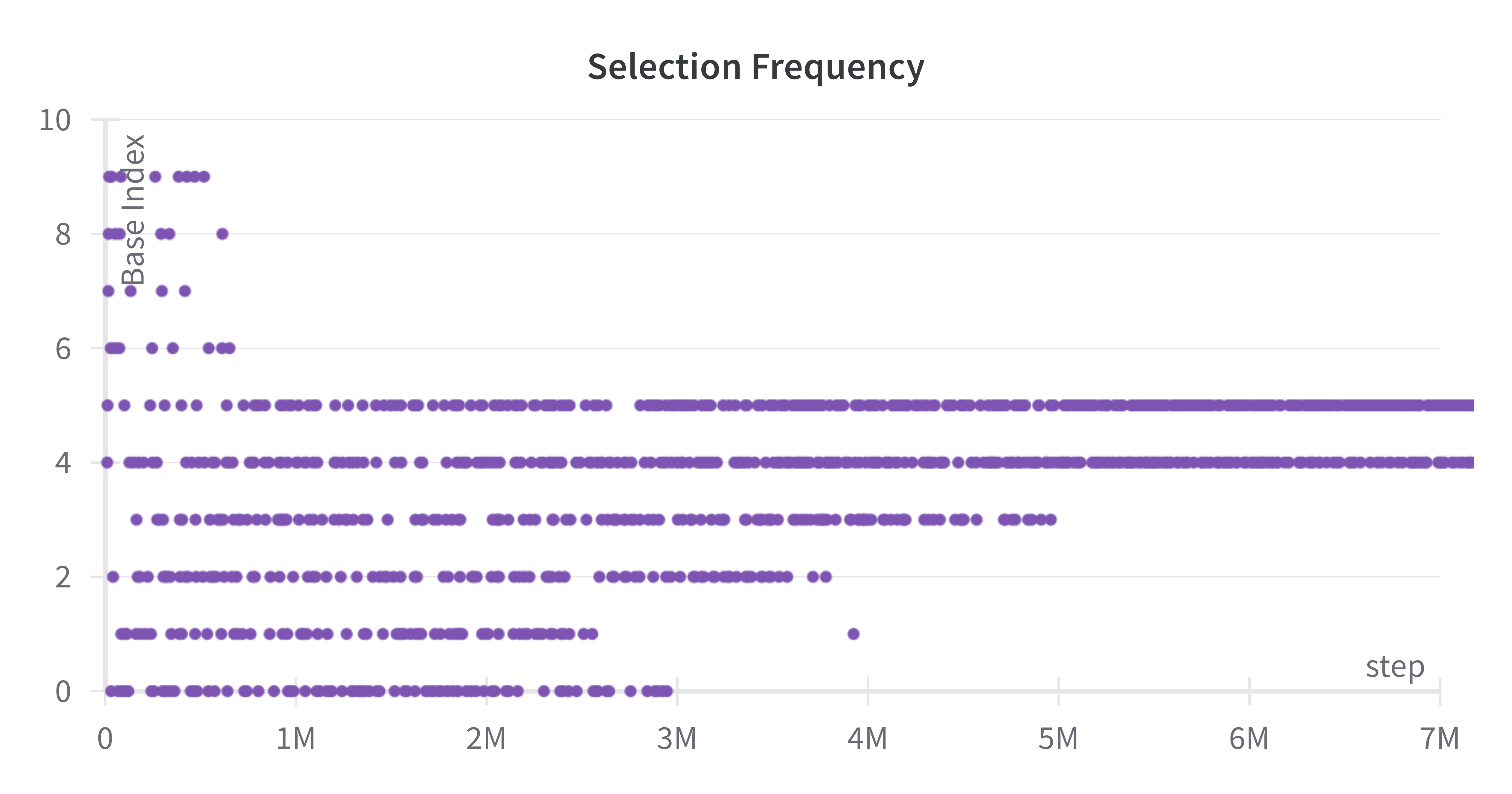}
  \caption{ED$^2$RB}
  \label{fig:4}
\end{subfigure}\hfil 
\begin{subfigure}{0.33\textwidth}
  \includegraphics[width=\textwidth]{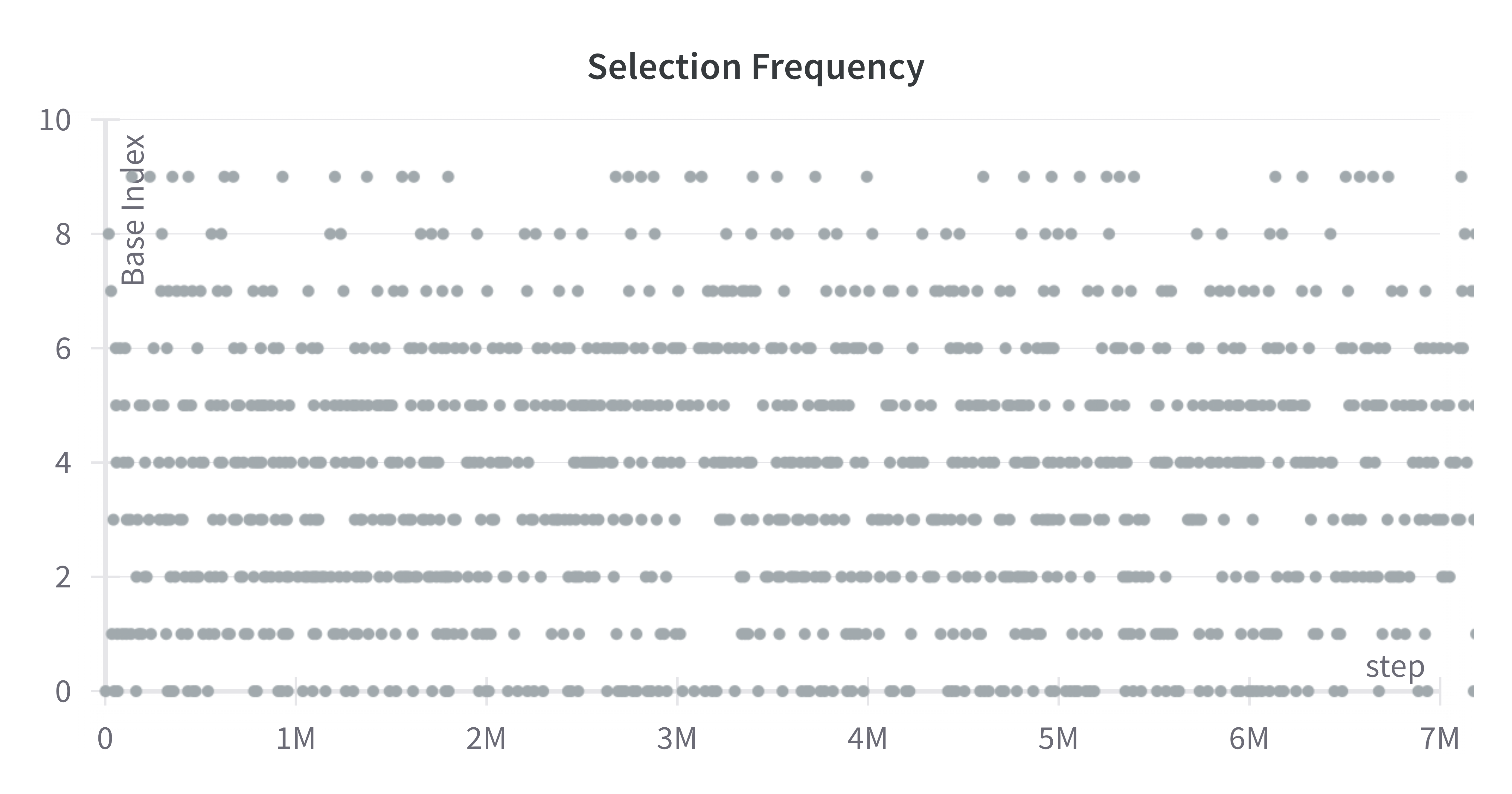}
  \caption{Classic Balancing}
  \label{fig:5}
\end{subfigure}\hfil 

\medskip

\begin{subfigure}{0.33\textwidth}
  \includegraphics[width=\textwidth]{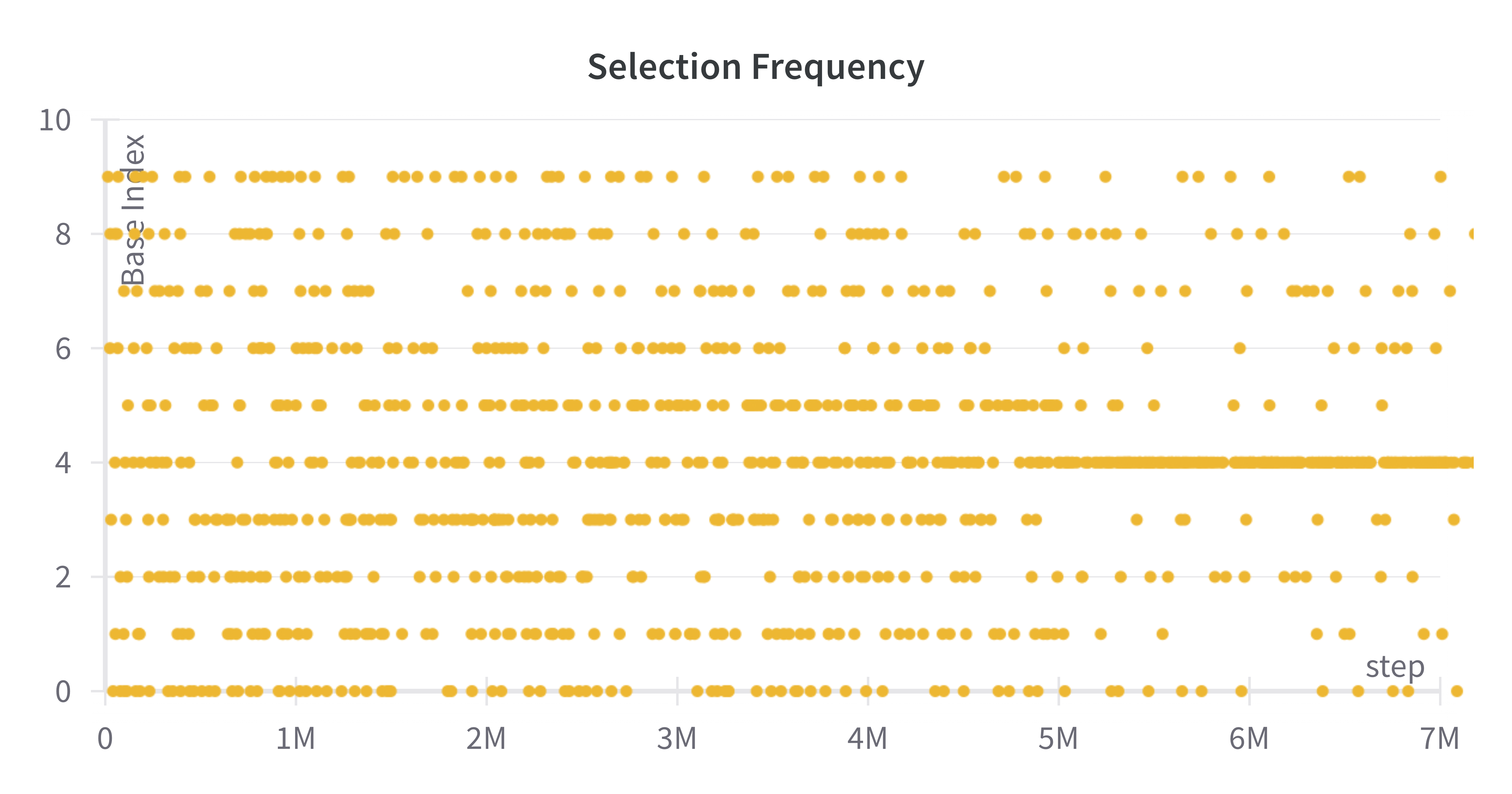}
  \caption{Corral}
  \label{fig:6}
\end{subfigure}\hfil 
\begin{subfigure}{0.33\textwidth}
  \includegraphics[width=\textwidth]{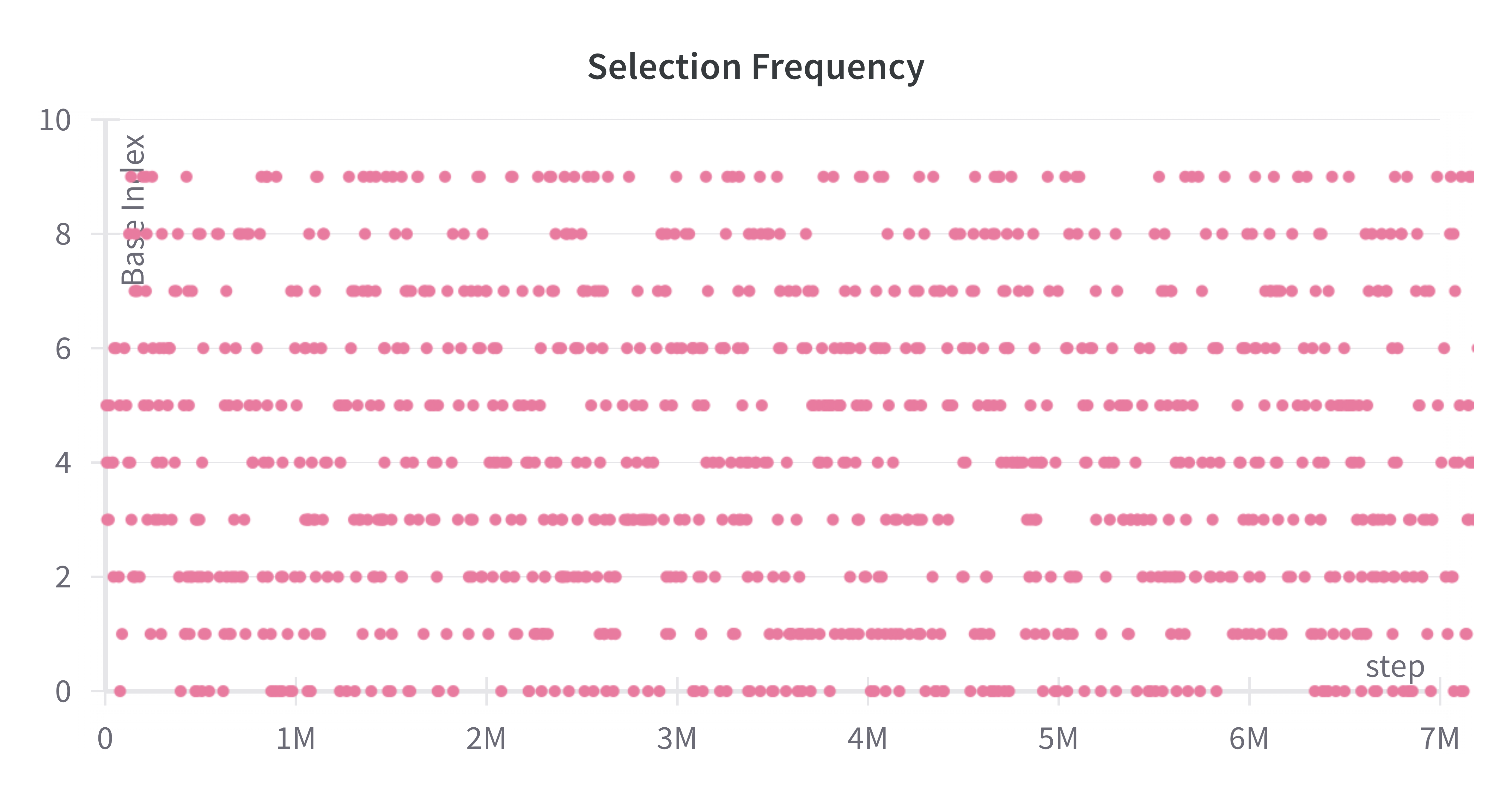}
  \caption{EXP3}
  \label{fig:7}
\end{subfigure}\hfil 
\begin{subfigure}{0.33\textwidth}
  \includegraphics[width=\textwidth]{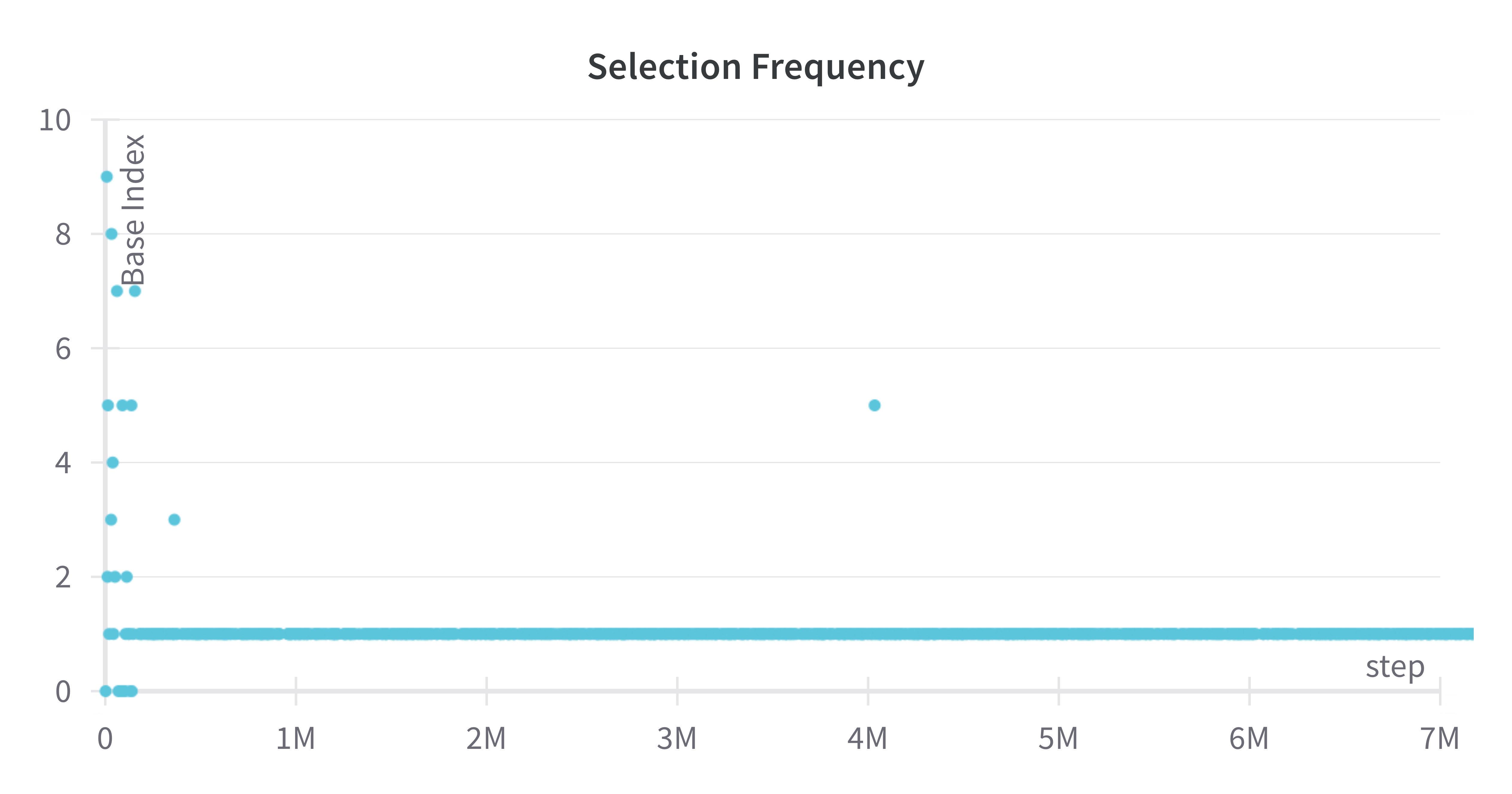}
  \caption{UCB}
  \label{fig:8}
\end{subfigure}

\caption{Selection frequency of each base learner for Learning Rate-Free PPO on Humanoid environment. The y-axis indicates the base learner's index, and the x-axis indicates the timestep. Each (x,y) point shows that base learner y was selected and played by meta learner at time x.}
\label{fig:fig2}
\end{figure}

To better understand the Model Selection strategies, we plotted the selection frequency of each base learner for all of the model selection strategies. \hyperref[fig:fig2]{Figure 2} represents our main findings. Each $(x,y)$ point in the plot shows that at time $x$ meta learner has selected base learner $y$. We can see that D$^3$RB and ED$^2$RB meta learners have learned to detect the best-performing base learners and played them more often, resulting in more advancement in the reward curve. \hyperref[fig:fig2]{Figure 2 (c, d, e)} show that meta learners with Classic Balancing, Corral, and EXP3 Strategies are selecting all the base learners quite often and therefore have an average performance in the task. \hyperref[fig:fig2]{Figure 2 (d)} also explains the slightly better performance of Corral, as this meta learner has learned to detect one of the optimal base learners at the end.

One major drawback of applying standard MAB algorithms such as UCB in non-stationary domains like RL can be seen in these plots. A specific choice of learning rate might achieve high rewards in the early stages of learning and not perform well later on. As seen in \hyperref[fig:fig2]{Figure 2 (f)} the base learner with index 1 was selected by UCB meta learner, since it was the best choice of learning rate in the early stages of learning. \hyperref[fig:fig1]{Figure 1} shows that this learning rate was performing well in the early stages of learning. Later on, this base learner was no longer an optimal learning rate but UCB couldn't adapt and continued to select that base learner. Algorithms like UCB are not able to distinguish these non-stationary changes and therefore are not suitable strategies for learning rate selection in reinforcement learning. The same challenge casts to common hyperparameter tuning techniques such as Bayesian Optimization, where the objective function is assumed to be stationary.

\begin{figure}[htb]
    \centering 

\begin{subfigure}{0.5\textwidth}
  \includegraphics[width=\textwidth]{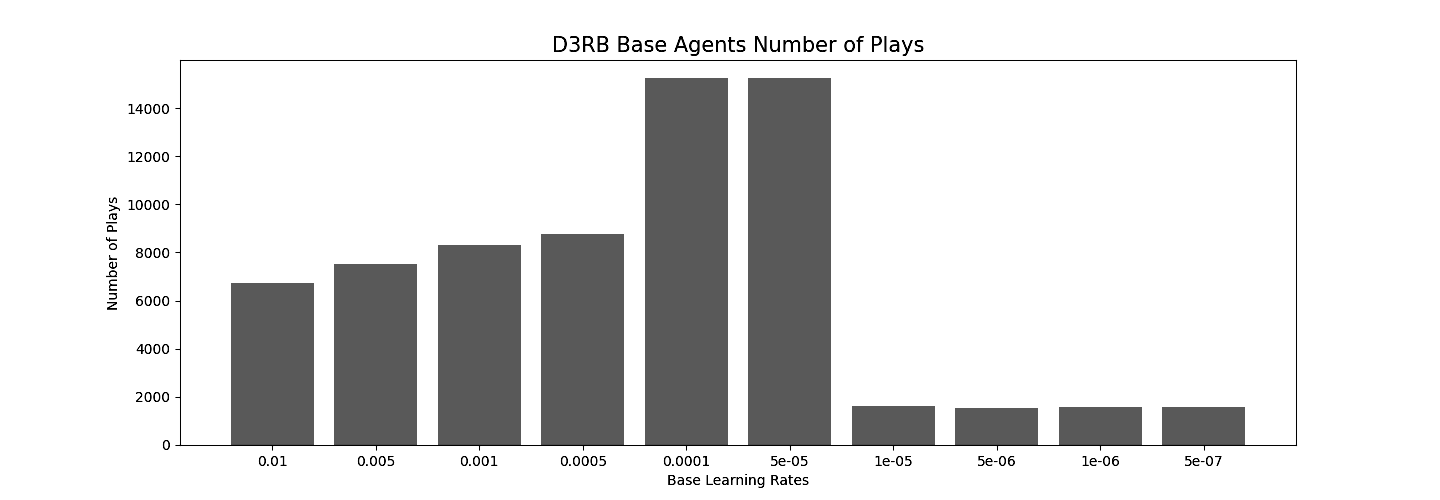}
  \label{fig:4}
\end{subfigure}\hfil 
\begin{subfigure}{0.5\textwidth}
  \includegraphics[width=\textwidth]{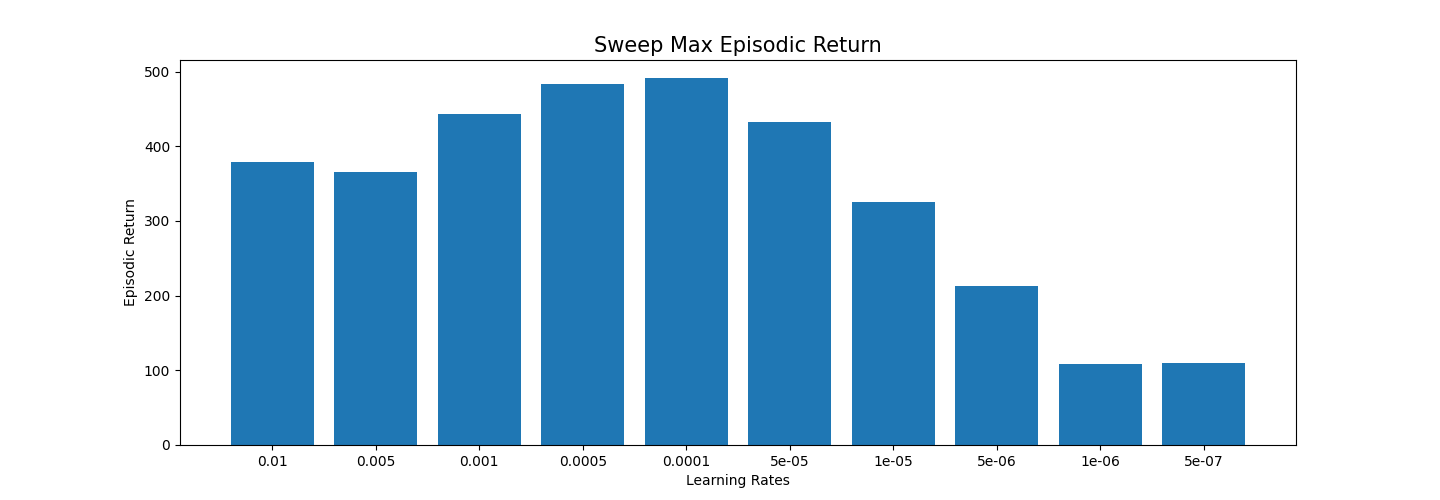}
  \label{fig:5}
\end{subfigure}\hfil 

\caption{\textbf{(Left)} Number of times D$^3$RB has played each learning rate. \textbf{(Right)} Maximum reward of PPO agents initiated with the same set of learning rates. We can see that D$^3$RB is playing base learners with higher rewards more frequently than base learners with suboptimal performance.}
\label{fig:fig3}
\end{figure}

Additionally, we initiate ten independent PPO agents with the same set of learning rates that we input to the model selection counterparts and run each agent for $\sim \frac{1}{10}$ fraction of total episodes in model selection experiments. \hyperref[fig:fig3]{Figure 3} demonstrates the results of this comparison. \hyperref[fig:fig3]{Figure 3(a)} shows the number of episodes that the meta learner with D$^3$RB strategy has selected each base agent throughout the training. \hyperref[fig:fig3]{Figure 3(b)} shows the maximum episodic return achieved by PPO agents initiated with the same learning rates. We can see that D$^3$RB strategy for learning rate-free PPO has learned to select the agents with higher reward (and lower regret) more frequently, and dedicate less sample and compute to suboptimal bases. In fact, through regret balancing a linearly suboptimal base will not be selected for more than $\sqrt{N}$ rounds, where $N$ is the total number of episodes. Check \hyperref[AppendixA]{Appendix A} for more theoretical details.

\section{Related Work}

Hyperparameter tuning rises as a crucial step in machine learning problems and various works have proposed methods for hyperparameter tuning since the early days of machine learning \citep{kohavi1995automatic, bergstra2011algorithms}. Random Search \citep{bergstra2012random} aimed to improve the naive grid search by considering a random subset of all possible hyperparameters, instead of the full grid. Bayesian optimization (BO)\citep{brochu2010tutorial, wu2019hyperparameter, ru2020bayesian} is a black-box optimization method that advanced the sample efficiency of naive approaches drastically. Despite being effective, BO requires multiple training runs to detect a good configuration of the problem, which can be deficient when combined with RL. This motivates the studying of dynamic scheduling that can be achieved in a single run. Several papers have studied learning rate scheduling, and learning rate-free learning \citep{defazio2023and, khaled2024tuning, mishchenko2023prodigy}, where they analyze the problem through the lens of convex optimization. 

Deploying Bandit algorithms for hyperparameter tuning is not a new idea. Hyperband \citep{li2018hyperband} formulates hyperparameter optimization as an infinite-arm bandit problem that tends to adaptively dedicate resources to the better-performing configurations and stop the ones with poor performance. In a sequential domain like RL, if one configuration is performing poorly at the early stages of learning but is optimal at the later stages, hyperband will not be able to detect it. Among the available literature, the closest to our work is \citep{parker2020provably} which uses a population-based bandit method for online hyperparameter optimization, where they fit a Gaussian Process model to select the hyperparameter at each step. Replacing the latter with methods that enjoy model selection guarantees is an interesting future direction.

\section{Conclusion and Future Work}
We proposed a model selection framework for learning rate-free reinforcement learning and demonstrated its effectiveness using six model selection strategies. Our experiments showed that the data-driven regret balancing method, D$^3$RB, and ED$^2$RB generally serve as good model selection strategies for learning rate-free reinforcement learning, consistently performing well across our tests. In contrast, bandit strategies appeared to be insufficient as meta-learners for PPO base agents. 

There are several possible extensions to this work. The span of hyperparameter optimization with model selection techniques is not limited to the learning rate. Applying model selection methods for tuning a set of different hyperparameters is an interesting direction that requires sample-efficient algorithms that can be deployed in RL. Studying the effect of sharing data across the base agents is another interesting direction that can further improve the efficiency and generalizability of the framework. 


\bibliography{main}
\bibliographystyle{rlc}

\appendix

\section*{Appendix}


\subsection*{A1. Theoretical Remarks on Regret Balancing}
Regret balancing strategies strive to maintain the regret of the different algorithms. Typically it is assumed the optimal algorithm's regret scales as $d_\star\sqrt{t}$. In contrast, the regret of a linearly sub-optimal algorithm scales as $\Delta t$ for some constant $\Delta$. Without loss of generality let's call these two algorithms, Algorithm~1 and Algorithm~2. A regret balancing strategy ensures that at time $N$ the number of times Algorithm~1 and Algorithm~2 were played, $N_1$, and $N_2$ satisfy $d_\star\sqrt{N_1} \approx \Delta N_2$ thus implying that $N_2 \approx \frac{d_\star \sqrt{N_1}}{\Delta} = \mathcal{O}\left(\frac{d_\star\sqrt{N}}{\Delta}\right) $.

\subsection*{A2. Preliminaries of Reinforcement Learning}

Reinforcement learning is formalized as Markov Decision Process (MDP) $\langle S, A, R, P, \gamma \rangle$; where $S$ denotes the set of states, $A$ is the set of actions, $R: S \times A \rightarrow \mathbb{R}$ is the reward function, $P: S \times A \rightarrow [0,1]$ is the dynamic transition probabilities, and lastly $\gamma \in [0,1]$ is the discount factor. Here we consider episodic reinforcement learning with maximum horizon $T$ where the goal of the agent is to learn the (near) optimal policy  $\pi:S \rightarrow A$. The state-value function $V: S \rightarrow \mathbb{R}$ and action-value function $Q: S \times A \rightarrow \mathbb{R}$ with respect to policy $\pi$ are defined as

\begin{equation}
    V^{\pi}(s) =\mathbb{E} \biggl[\sum_{t=0}^{T} \gamma^t R(s_t,a_t) | s_0=s, s_t, a_t \biggr] 
\end{equation}
\begin{equation}
    Q^{\pi}(s, a) = R(s, a) + \gamma \: \mathbb{E}_{s^{\prime} \sim P(s,a)}\biggl[ V^{\pi}(s^{\prime})\biggr]
\end{equation}

The policy $\pi$ is commonly parameterized by the set of parameters $\theta$, and is denoted as $\pi_{\theta}$.  Two of the predominant approaches for learning the (near) optimal policy in reinforcement learning are policy optimization and Q-learning. Policy optimization starts with an initial policy and in each episode updates the parameters 
by taking gradient steps toward maximizing the episodic return. Denote learning rate as $\alpha \in \mathbb{R}$, a common update rule in policy optimization methods is
\begin{equation}
    \theta \leftarrow \theta + \alpha \: \mathbb{E} \biggl[ \sum_{t=0}^{T} \nabla_{\theta} \log \pi_{\theta} (s_t, a_t) (Q^{\pi_{\theta}}(s_t, a_t) - V^{\pi_{\theta}}(s_t)) \biggr]
\end{equation}

Q-learning uses the temporal differences method to update the parameters of $Q^{\pi_{\theta}}$. A common update rule is
\begin{equation}
    \theta \leftarrow \theta + \alpha  \: \mathbb{E}_{s,a,s^{\prime}, r \sim D} \biggl[\nabla_{\theta} (r + \gamma \: \max_{a^{\prime} \in A} Q^{\pi_{\bar\theta}}(s^{\prime}, a^{\prime}) - Q^{\pi_{\theta}}(s, a))^2 \biggr]
\end{equation}
where $D$ is the experience replay buffer and $\bar{\theta}$ is a frozen parameter set named target parameter. Proximal Policy Optimization (PPO) \citep{schulman2017proximal} and Deep Q-Networks (DQN) \citep{mnih2015human} follow the first and second approaches, respectively.

\subsection*{A3. Learning Rate-Free Reinforcement Learning}

\begin{algorithm}[!ht]
\DontPrintSemicolon
  
  \KwInput{MDP $\langle S, A, R, P, \gamma \rangle$, Model Selection Interface $\mathfrak{M}$}
  
  \;
  \tcp{reinforcement learning loop over episode}
  \For{$n = 1, 2, ..., N$}
    {
        \tcp{Select the base agent}
        $i, \pi^i_{\theta}, \alpha^i = \mathfrak{M}.sample() $ \;

        \tcp{Collect trajectories with selected base agent}
        \For{$t = 1, 2, ..., T$}    
        {
            $a \sim \pi^i_{\theta}$ \;
            $r, s^{\prime} \xleftarrow[]{P, R} s, a$ \;
            $R[t] \leftarrow r$ \;
        }
        \tcp{Update parameters with selected learning rate}
        \If{Policy Optimization}
            {
            $\theta \leftarrow \theta + \alpha^i \: \mathbb{E} \biggl[ \sum_{t=0}^{T} \nabla_{\theta} \log \pi^i_{\theta} (s_t, a_t) (Q^{\pi^i_{\theta}}(s_t, a_t) - V^{\pi^i_{\theta}}(s_t)) \biggr]$
            }
        \If{Q-Learning}
            {
            $\theta \leftarrow \theta + \alpha^i  \: \mathbb{E}_{s,a,s^{\prime}, r \sim D} \biggl[\nabla_{\theta} (r + \gamma \: \max_{a^{\prime} \in A} Q^{\pi^i_{\bar\theta}}(s^{\prime}, a^{\prime}) - Q^{\pi^i_{\theta}}(s, a))^2 \biggr]$
            }
        \tcp{Update the meta learner}
        $\mathfrak{M}.update(i, R[1:T])$  \;
    }

\label{alg2}
\caption{Learning Rate-Free Reinforcement Learning with Model Selection}
\end{algorithm}

\subsection*{B. Codes}
 All codes and implementations of the learning rate-free RL algorithms are available at \url{github.com/AidaAfshar/Learning-Rate-Free-Reinforcement-Learning}. We use cleanRL library \citep{huang2022cleanrl} for the implementation of RL algorithms. The model selection strategies were originally implemented at \url{https://github.com/pacchiano/modelselection}. 






\subsection*{C. Model Selection Algorithms Pseudocodes}
\label{AppendixC}
We provide the Model Selection Algorithms in hyperparparemeter tuning interface in this section. To avoid including all the theoretical details, there might be a slight abuse of notation in the pseudocodes. We encourage the reader to check the details with the original paper.

Denote the number of times that the base agent $i$ was played up to this time as $n^i$. Denote the regret coefficient of base learner $i$ as  $d^i$, and the total reward accumulated by base learner $i$ up to this time by $u^i$.

\subsubsection*{D$^3$RB}
 Doubling Data Driven Regret Balancing (D$^3$RB) \citep{dann2024data} tries to maintain and equal empirical regret for all the base agents. Denote the balancing potential of base agent $i$ as $\Psi^i = d^i\sqrt{n^i}$. The D$^3$RB algorithm for learning rate-free RL works as follows,

\begin{algorithm}[H]
\SetKwInput{KwInput}{Input}                
\SetKwInput{KwOutput}{Output}              
\DontPrintSemicolon
  
  \KwInput{$m$, $\beta$, $\Psi$, $\delta$}

  \SetKwFunction{FMain}{Main}
  \SetKwFunction{FSum}{sample}
  \SetKwFunction{FSub}{update}
 
  \SetKwProg{Fn}{Function}{:}{}
  \Fn{\FSum{}}{
        \tcp{Sample base index}
        $i = \argmin_j \Psi_j$\;
        $\pi_i, \alpha_i \leftarrow \beta_i$ \;
        \KwRet $i, \pi_i, \alpha_i$\;
  }

  \SetKwProg{Fn}{Function}{:}{}
  \Fn{\FSub{$i$, $R[1:T]$}}{ 
      $R_{norm} \leftarrow normalize (R[1:T])$ \;
      \tcp{Update Statistics}
        $u^i = u^i+R_{norm}$ \;
        $n^i = n^i + 1$ \;
    \tcp{Perform miss-specification test}
        $ \frac{u^i}{n^i} + \frac{d^i \sqrt{n^i}}{n^i} + c \sqrt{ln \frac{\frac{m ln n^i}{\delta}}{n^i}} \leq max_j \frac{u^j}{n^j} - c \sqrt{ln \frac{\frac{M ln n^j}{\delta}}{n^j}}$ \;
    \tcp{If test triggered double regret coefficient for base i}
        $d^i \leftarrow 2d^i$ \;
    \tcp{Update balancing potential}
        $\Psi^i = d^i\sqrt{n^i}$ \;
    
  }
\label{alg3}
\caption{D$^3$RB}
\end{algorithm}

\subsubsection*{ED$^2$RB}
 Estimating Data Driven Regret Balancing (ED$^2$RB) \citep{dann2024data} is similar to D$3$RB, though it tries to directly estimate the regret coefficients.

\begin{algorithm}[H]
\SetKwInput{KwInput}{Input}                
\SetKwInput{KwOutput}{Output}              
\DontPrintSemicolon
  
  \KwInput{$m$, $\beta$, $\Psi$, $\delta$}

  \SetKwFunction{FMain}{Main}
  \SetKwFunction{FSum}{sample}
  \SetKwFunction{FSub}{update}
 
  \SetKwProg{Fn}{Function}{:}{}
  \Fn{\FSum{}}{
        \tcp{Sample base index}
        $i = \argmin_j \Psi_j$\;
        $\pi_i, \alpha_i \leftarrow \beta_i$ \;
        \KwRet $i, \pi_i, \alpha_i$\;
  }

  \SetKwProg{Fn}{Function}{:}{}
  \Fn{\FSub{$i$, $R[1:T]$}}{ 
      $R_{norm} \leftarrow normalize (R[1:T])$ \;
      \tcp{Update Statistics}
        $u^i = u^i+R_{norm}$ \;
        $n^i = n^i + 1$ \;
    \tcp{Estimate active regret coefficient}
        $d^i = max{d_{min}, \sqrt{n_t^{i_t}}(max_j \frac{u^j}{n^j} - c \sqrt{ln \frac{\frac{M ln n^j}{\delta}}{n^j}} - c \sqrt{ln \frac{\frac{m ln n^i}{\delta}}{n^i}} - \frac{u^i}{n^i})}
        $ \;
    \tcp{Update balancing potential}
        $\Psi^i = clip(d^i\sqrt{n^i}, \Psi^i, 2\Psi^i)$ \;
    
  }
\label{alg3}
\caption{ED$^2$RB}
\end{algorithm}

\subsubsection*{Classic Balancing}
The Classic Regret Balancing Algorithm \citep{pacchiano2020regret} starts with the full set of base agents $\beta = [\beta_1, ..., \beta_m]$, at each round the algorithm performs miss-specification on each of the base agents and eliminates the miss-specified one. Denote $\Psi^j$ as empirical regret upper bound of base agent $j$.

\begin{algorithm}[H]
\SetKwInput{KwInput}{Input}                
\SetKwInput{KwOutput}{Output}              
\DontPrintSemicolon
  
  \KwInput{$m$, $\beta$, $\Psi$, $\delta$}

  \SetKwFunction{FMain}{Main}
  \SetKwFunction{FSum}{sample}
  \SetKwFunction{FSub}{update}
 
  \SetKwProg{Fn}{Function}{:}{}
  \Fn{\FSum{}}{
        \tcp{Sample Base index}
        $i = \argmin_j \Psi_j$\;
        $\pi_i, \alpha_i \leftarrow \beta_i$ \;
        \KwRet $i, \pi_i, \alpha_i$\;
  }

  \SetKwProg{Fn}{Function}{:}{}
  \Fn{\FSub{$i$, $R[1:T]$}}{
      $R_{norm} \leftarrow normalize (R[1:T])$ \;
      
        \tcp{Update statistics}
        $u^i = u^i+R_{norm}$ \;
        $n^i = n^i + 1$ \;

        \tcp{Perform miss-specification test for all the remaining base agents}
        \For{$\beta_k\in \beta$}
        {$ \frac{u^k}{n^k} + \frac{d^k \sqrt{n^k}}{n^i} + c \sqrt{ln \frac{\frac{m ln n^k}{\delta}}{n^k}} \leq max_j \frac{u^j}{n^j} - c \sqrt{ln \frac{\frac{M ln n^j}{\delta}}{n^j}}$ \;
        \If{miss-specified}{
            $\beta \leftarrow \beta / \{\beta_{k}\}$ \;
        } \;
        }

  }
\label{alg4}
\caption{Classic Balancing}
\end{algorithm}

\subsubsection*{EXP3}
Exponential-weight algorithm for exploration and exploitation (EXP3) learns a probability distribution $\Psi^i = \frac{exp(S^i)}{\sum_{j=1}^{m} exp(S^j)}$ over base learners, where $S^i$ is a total estimated reward of base agent $i$ up to this round.

\begin{algorithm}[H]
\SetKwInput{KwInput}{Input}                
\SetKwInput{KwOutput}{Output}              
\DontPrintSemicolon
  
  \KwInput{$m$, $\beta$, $\Psi$, $\delta$}

  \SetKwFunction{FMain}{Main}
  \SetKwFunction{FSum}{sample}
  \SetKwFunction{FSub}{update}
 
  \;
  \SetKwProg{Fn}{Function}{:}{}
  \Fn{\FSum{}}{
        \tcp{Sample Base index}
        $i = \argmax_j \Psi_j$\;
        
        $\pi_i, \alpha_i \leftarrow \beta_i$ \;
        \;
        \KwRet $i, \pi_i, \alpha_i$\;
  }
  \SetKwProg{Fn}{Function}{:}{}
  \Fn{\FSub{$i$, $R[1:T]$}}{ \;
      $R_{norm} \leftarrow normalize (R[1:T])$ \;
      \tcp{Update statistics}
        
        \For{$j\in {1, ..., m}$} 
          {$S^j = S^j+ 1 - \frac{\mathbb{I}\{j = i\}(1-R_{norm})}{\Psi^i}$}
        \;
    \tcp{Update Distribution}
        $\Psi^i = \frac{exp(S^i)}{\sum_{j=1}^{m} exp(S^j)}$ \;
    
  }
\label{alg5}
\caption{EXP3}
\end{algorithm}

\subsubsection*{Corral}
Corral \citep{agarwal2017corralling} learns a distribution $\Psi$ over base agents and update it according to LOG-BARRIER-OMD algorithm. We skip the algorithmic details and refer to the updating rule mentioned in the original paper as Corral-Update.

\begin{algorithm}[H]
\SetKwInput{KwInput}{Input}                
\SetKwInput{KwOutput}{Output}              
\DontPrintSemicolon
  
  \KwInput{$m$, $\beta$, $\Psi$}

  \SetKwFunction{FMain}{Main}
  \SetKwFunction{FSum}{sample}
  \SetKwFunction{FSub}{update}
 
  \;
  \SetKwProg{Fn}{Function}{:}{}
  \Fn{\FSum{}}{
        \tcp{Sample base index}
        $i \sim \Psi$\;
        
        $\pi_i, \alpha_i \leftarrow \beta_i$ \;
        \;
        \KwRet $i, \pi_i, \alpha_i$\;
  }

  \SetKwProg{Fn}{Function}{:}{}
  \Fn{\FSub{$i$, $R[1:T]$}}{ \;
      $R_{norm} \leftarrow normalize (R[1:T])$ \;
      \tcp{Update according to Corral}
        $\Psi^j \leftarrow \text{Corral-Update}(R_{norm})$ \;
    
  }
\label{alg6}
\caption{Corral}
\end{algorithm}

\subsubsection*{UCB}
The Upper Confidence Bound algorithm (UCB) maintains an optimistic estimate of the mean for each arm \citep{lattimore2020bandit}. Denote $\Psi^i$ as the upper confidence bound of arm $i$. The UCB algorithm  for learning rate-free RL works as follows,

\begin{algorithm}[H]
\SetKwInput{KwInput}{Input}                
\SetKwInput{KwOutput}{Output}              
\DontPrintSemicolon
  
  \KwInput{$m$, $\beta$, $\Psi$, $\delta$}

  \SetKwFunction{FMain}{Main}
  \SetKwFunction{FSum}{sample}
  \SetKwFunction{FSub}{update}
 
  \SetKwProg{Fn}{Function}{:}{}
  \Fn{\FSum{}}{
        \;
        \tcp{Sample base index}
        $i = \argmax_j \Psi_j$\;
        
        $\pi_i, \alpha_i \leftarrow \beta_i$ \;
        \;
        \KwRet $i, \pi_i, \alpha_i$\;
  }
  
  \SetKwProg{Fn}{Function}{:}{}
  \Fn{\FSub{$i$, $R[1:T]$}}{ \;
      $R_{norm} \leftarrow normalize (R[1:T])$ \;
      \tcp{Update statistics}
        $u^i = u^i+R_{norm}$ \;
        $n^i = n^i + 1$ \;
        $\mu^i = \frac{u^i}{n^i}$
        
    \tcp{Update Upper Confidence Bounds}
        $\Psi^i = UCB^i (\delta) =  \mu^i + \sqrt{\frac{2 log(1 / \delta)}{n^i}}$ \;
    
  }
\label{alg7}
\caption{UCB}
\end{algorithm}

\end{document}